\documentclass[conference]{IEEEtran}
\IEEEoverridecommandlockouts
\usepackage{cite}
\usepackage{amsmath,amssymb,amsfonts}
\usepackage{algorithmic}
\usepackage{graphicx}
\usepackage{textcomp}
\usepackage{xcolor}
\usepackage[T1]{fontenc}
\usepackage{float}
\usepackage{newtxmath}
\usepackage{multirow}
\usepackage{amssymb}
\usepackage{changes}
\usepackage{booktabs} 
\usepackage{caption} 
\def\BibTeX{{\rm B\kern-.05em{\sc i\kern-.025em b}\kern-.08em
    T\kern-.1667em\lower.7ex\hbox{E}\kern-.125emX}}
\begin{document}
\title{GRAINRec: Graph and Attention Integrated Approach for Real-Time Session-Based Item Recommendations\\
}

\makeatletter
\newcommand{\linebreakand}{%
  \end{@IEEEauthorhalign}
  \hfill\mbox{}\par
  \mbox{}\hfill\begin{@IEEEauthorhalign}
}
\makeatother

\author{
    \IEEEauthorblockN{Bhavtosh Rath}
    \IEEEauthorblockA{Data Sciences, Target Corporation \\
    Brooklyn Park, MN, USA \\
    bhavtosh.rath@target.com}
    \and
    \IEEEauthorblockN{Pushkar Chennu}
    \IEEEauthorblockA{Data Sciences, Target Corporation \\
    Brooklyn Park, MN, USA \\
    pushkar.chennu@target.com}
    \and
    \IEEEauthorblockN{David Relyea}
    \IEEEauthorblockA{Data Sciences, Target Corporation \\
    Brooklyn Park, MN, USA \\
    david.relyea@target.com}
    \and
    \linebreakand
        \IEEEauthorblockN{Prathyusha Kanmanth Reddy}
    \IEEEauthorblockA{Data Sciences, Target Corporation \\
    Brooklyn Park, MN, USA \\
    prathyusha.kanmanthreddy@target.com}
    \and
    \IEEEauthorblockN{Amit Pande}
    \IEEEauthorblockA{Data Sciences, Target Corporation \\
    Brooklyn Park, MN, USA \\
    amit.pande@target.com}
    \and

}


\maketitle
\begin{abstract}
Recent advancements in session-based recommendation models using deep learning techniques have demonstrated significant performance improvements. 
While they can enhance model sophistication and improve the relevance of recommendations, they also make it challenging to implement a scalable real-time solution. To addressing this challenge, we propose GRAINRec- a \underline{Gr}aph and \underline{A}ttention \underline{In}tegrated session-based \underline{rec}ommendation model that generates recommendations in real-time. Our scope of work is item recommendations in online retail where a session is defined as an ordered sequence of digital guest actions, such as page views or adds to cart. The proposed model generates recommendations by considering the importance of all items in the session together, letting us predict relevant recommendations dynamically as the session evolves. We also propose a heuristic approach to implement real-time inferencing that meets Target platform's service level agreement (SLA). The proposed architecture lets us predict relevant recommendations dynamically as the session evolves, rather than relying on pre-computed recommendations for each item. Evaluation results of the proposed model show an average improvement of 1.5\% across all offline evaluation metrics. A/B tests done over a 2 week duration showed an increase of 10\% in click through rate and 9\% increase in attributable demand. Extensive ablation studies are also done to understand our model performance for different parameters.
\end{abstract}


\section{Introduction}
Recommender systems play an important role in retail guest experiences by predicting and displaying products that guests would be most interested in purchasing. As per a report by BusinessWire\footnote{https://www.businesswire.com/news/home/20220530005180/en/Artificial-Intelligence-in-Retail-2022-Market-Research-Report---Global-Industry-Analysis-and-Growth-Forecast-to-2030---ResearchAndMarkets.com} by 2025, e-commerce sales are expected to reach \$7.3 trillion, which will drive the AI in retail market value to \$36,462.5 million by 2030 from an estimated \$1,714.3 million in 2021. It is thus a critical area of research, not just in academia but more importantly in industry. 
In this paper we propose and analyze the quality of recommendations for a session-based recommendation model with real time inferencing capability. 

To explain what a session-based recommendation model is, we first define a \textit{session}. A \textit{session} is a sequence of  ordered items in close temporal proximity. In the context of item recommendations in online retail, examples of sessions are items browsed, purchased or added to cart having a temporal proximity of usually a few minutes. Session-based recommendations face two major challenges over traditional single item based recommender systems: \\
\textit{1)} While traditional models provide recommendations in the context of the current item being considered by a user, a session based model generates recommendations considering the overall context of all items in the session, which is comparatively more challenging to understand. Suppose a guest intends to buy three items in a session. When milk is added as the first item, traditional recommendations are items related to milk like almond milk, butter, and yogurt. If the user adds eggs next, traditional models tend to recommend items related to eggs (like egg whites, brown eggs, 12-count eggs, 18-count eggs). However, GRAINRec considers the combined context (milk and eggs) to recommend breakfast items like bread, cereal, and coffee. If sugar gets added as the third item, existing models might recommend items related to sugar like stevia sweetener, granulated sugar, and organic sugar. GRAINRec would consider the basket context of all three items and recommend baking items - like cake flour, frosting, and baking powder.\\
\textit{2)} A major challenge for session based models is real-time inferencing. For single item based recommender models, collaborative filtering-based or item embedding-based recommendations can be generated offline and used in a lookup in real time. In the previous example, for a retailer with a catalog of say 1 Million items, a session of length just N = 3 would require a $10^{12}$ sized lookup table!  These kinds of dynamic recommendations cannot be pre-computed using existing models.

The contributions of the paper can be summarized as follows: 

\textbf{1.} We propose GRAINRec which builds upon the foundation of the LESSR model \cite{chen2020handling} while incorporating enhancements (detailed in Section V) critical to achieving both high performance and scalability in a high rate real-time framework.

\textbf{2.} Retail sessions evolve dynamically so the generation of item recommendations needs to happen in real time. This paper addresses this concern via a nearest neighbor matrix approach.

\textbf{3.} The paper presents comprehensive offline and online evaluation results, along with ablation studies that demonstrate GRAINRec's effectiveness. Additionally, we provide a detailed explanation of the model parameters and the inference setup used to deploy the training and inference architecture on Target's platform.

\section{Related Work}
\subsection{Session-based recommender systems}
Session-based item recommendation models have garnered significant attention in recent years due to their ability to provide personalized recommendations based on item relationships. They model dynamic user preferences and provide recommendations that are sensitive to the evolution of session context. There have been many papers suggesting different session-based recommendation models applied to different use cases. Wang et. al. \cite{wang2021survey} performed a comprehensive survey in this regard. In the domain of news recommendations Song et. al. \cite{song2016multi} proposed a session-based model to optimize for recommendations that are more focused on the freshness of the content while also modeling long term preferences with regards to topics. Shen et. al. \cite{shen2022session} proposed a recency-regularized neural attentive framework that uses users' sequential commenting behavior to recommend news. For music recommendation Wang et. al. \cite{wang2016learning} proposed a context-based model that mines knowledge from public opinion such as comments, music review, or social tags to recommend music. Bao et. al. \cite{bao2012location} presented a session-based location recommendation system that leverages sparse geo-social data. It combines user preferences and geographic information to suggest locations. Wang et. al. \cite{wang2021time} presented a dynamic time-aware attention model for recommending points of interest (POIs) by analyzing users' check-in sessions, capturing temporal patterns and user preferences. For video recommendations Beutel et. al. \cite{beutel2018latent} proposed a recurrent neural network-based approach on sequences of watches on YouTube to integrate the session context.

\subsection{Session-based recommendations for items}
This subsection focuses only on papers proposed for item recommendations. Yap et. al. \cite{yap2012effective} proposed  a sequential pattern mining based next-items recommendation algorithm to predict users' next accesses by identifying frequent patterns. The paper introduces a personalized framework using a competence score measure to improve the accuracy of recommendations for individual users. Hu et. al. \cite{hu2020modeling} proposed a KNN-based model that captures personalized item frequency information. Le et. al. \cite{le2016modeling} proposed a Markov chain-based model that introduces generative modeling that incorporates dynamic user-biased emission and context-biased transition, improving next-item prediction accuracy. Hidasi et. al.\cite{hidasi2015session} introduced GRU4Rec that used Gated Recurrent Units (GRUs) for session-based recommendations. GRU4Rec's ability to model complex user behavior sequences laid the foundation for subsequent research in session-based recommendations using deep learning. Li et. al.\cite{li2017neural} proposed NARM, a neural attentive model that incorporated an attention mechanism to focus on relevant parts of the session history, enhancing the recommendation accuracy. Kang and McAuley \cite{kang2018self} introduced SASRec, a self-attentive sequential recommendation model that employs the transformer architecture. By leveraging self-attention, SASRec captures long-range dependencies within a session more effectively than RNN-based approaches. Wang et. al. \cite{wang2020next} proposed a novel approach using hypergraphs for session-based recommendations. The Sequential Hypergraph Attention Network (SHAN) models user sessions as hypergraphs, where hyperedges connect multiple items. This structure allows the model to capture higher-order item co-occurrences and complex dependencies within sessions. SHAN's attention mechanism further enhances its ability to focus on the most relevant parts of the session, leading to improved recommendation performance. Feng et. al.\cite{tang2018personalized} introduced Caser, a Convolutional Sequence Embedding Recommendation model, which applies convolutional neural networks (CNNs) to model user behavior sequences. Caser captures both point-wise and union-level sequential patterns by treating the embedding matrix as an "image" and applying horizontal and vertical convolutional filters. The model showed significant improvement over traditional sequential recommendation methods. Zhang et. al. \cite{shi2018heterogeneous} proposed HINRec, a session-based recommendation model using Heterogeneous Information Network (HIN) embeddings. This model captures various types of interactions and relationships among items by representing the session data as a heterogeneous graph. The embeddings generated from this network effectively incorporate both structural and semantic information, resulting in improved accuracy. Liu et. al.\cite{liu2018stamp} proposed STAMP, a session-based recommendation model that integrates short-term attention and long-term memory. It utilizes an attention mechanism to capture users' short-term interests within a session, while a memory network stores long-term preferences. This dual approach allows STAMP to dynamically balance between recent interactions and long-standing preferences, achieving superior performance.

\subsection{Session-based recommender models on graph data}
This subsection highlights session based recommendation research on graph data. Gao et. al.\cite{gao2023survey} conducted a comprehensive review of the literature on graph neural network-based recommender systems. Wu et. al. \cite{wu2019session} proposed SR-GNN which models sessions as graphs where items are nodes and interactions are edges. The model utilizes Graph Neural Networks (GNNs) to capture complex item transitions within sessions, significantly improving recommendation accuracy.  Wang et. al.\cite{pan2020star} proposed a gated GNN model for session-based recommendations. The gated mechanism helps the model selectively focus on important parts of the session graph, enhancing the overall recommendation performance by capturing more nuanced user behaviors. Wang et. al.\cite{wang2019knowledge} introduced a knowledge-aware GNN that incorporates external knowledge graphs into the recommendation process. By integrating knowledge with session data and applying label smoothness regularization, the model achieves improved recommendation accuracy. Zhou et. al.\cite{li2023session} presented Temporal Graph Neural Networks (TGNN) for session-based recommendation. TGNN captures temporal dynamics within sessions by modeling time-aware item transitions. This approach enhances the model's ability to understand user behavior changes over time, leading to more accurate recommendations. Chen et. al.\cite{chen2020handling} proposed LESSR, a GNN and attention based approach this work borrows the model from. However their work does not propose a real-time inferencing solution.

\section{Problem Definition}
A session-based item recommendation model will recommend the next best item relevant to the context of item(s) in the session. Let  $I = \{i_1, i_2, . . . , i_n\}$ denote the set of all $n$ items in the model. A session  $ S_i= [s_{(i,1)}, s_{(i,2)} , . . . ,s_{(i,t)}]$ is an item sequence of length $t$ with items ordered with respect to time and $s_{(i,)} \in I$. The objective of the model is to predict the next item $s_{(i,t+1)}$. 

\section{Model Architecture}

\begin{figure*}[ht]
\centering
\includegraphics[width=\linewidth]{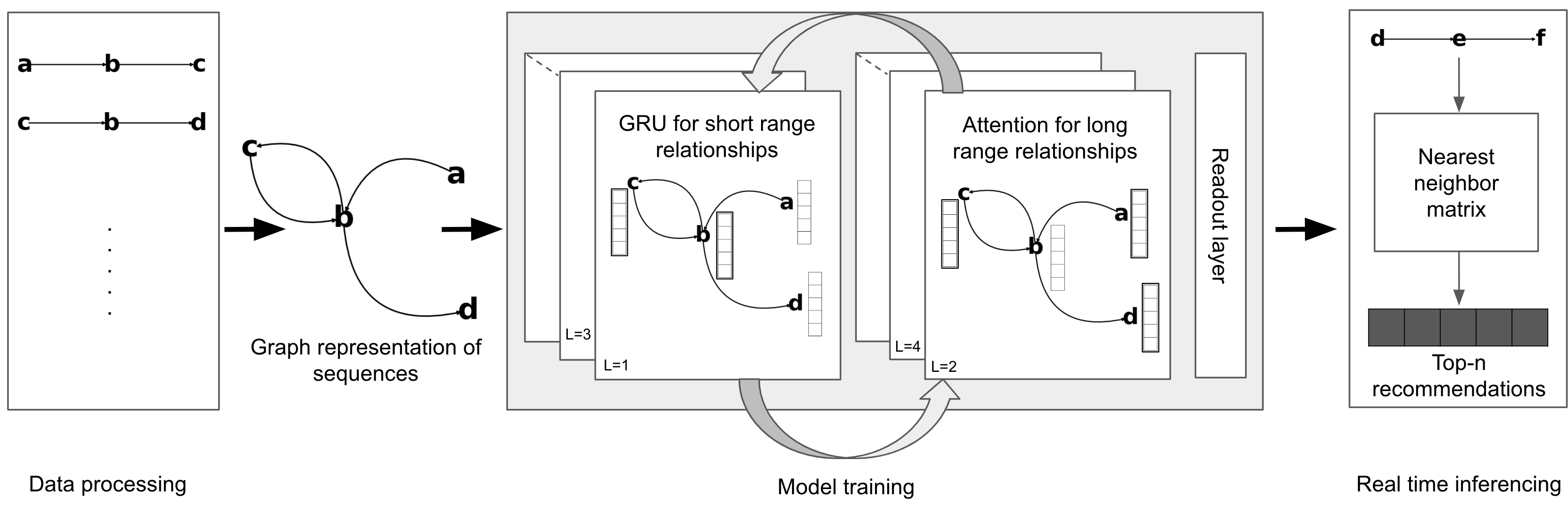}
\caption{Architecture of GRAINRec. Data processing generates ordered sequences of items interacted within a session, which are batched together as directed graphs. Model training consists of interchanging graph neural network and attention layers that generate item embeddings, followed by a readout layer that generates session embeddings. A nearest neighbor matrix is used to perform real-time inferencing.}
\label{fig:architecture}
\end{figure*}

GRAINRec's architecture can be divided into 3 modules: data processing, model training, and real-time inferencing. The architecture is shown in Figure~\ref{fig:architecture}. We explain about each module in detail below.

\subsection{Data Processing}
In the \textbf{data processing} phase we generate a temporally-ordered sequence of items added to a cart. Each sequence can be represented as an ordered graph. For example, if a guest adds three items $a$, $b$ and $c$ in that order then it can be represented as a directed graph ($a \rightarrow b \rightarrow c$), where the edge $a \rightarrow b$ means item $b$ was added to cart after item $a$. The graph does not contain self-loops ($a \rightarrow a$), but the same item $a$ can be added more than once in the sequence. During model training the sequences are batched, which means all sequences in the batch are merged to generate a single graph representation. An illustration is shown in Figure~\ref{fig:architecture}, where two ordered sequences ($a \rightarrow b \rightarrow c$ and $c\rightarrow b\rightarrow d$) are merged to generate a single graph representation.\\

\subsection{Model Training}
The proposed session-based recommendation architecture is a deep learning model that has two types of neural network layers: A Gated Recurrent Unit (GRU)-based graph neural network that captures short-range relationships in the sequences and an attention-based model that captures long-range relationships in the sequences. For simplicity we show in Figure~\ref{fig:architecture} that item \( c \) predominantly learns from item \( b \) and items \( a, d \) using GRU and attention mechanisms respectively. 
The model alternates between layer types; \textit{layer=1} is \textit{GNN}, \textit{layer=2} is \textit{GNN-Attention}, \textit{layer=3} is \textit{GNN-Attention-GNN}, and so on. This helps propagate the features captured by each kind of layer more effectively. We also apply a readout layer that calculates a session embedding by concatenating the attention-based aggregate of all ordered items in the session (called the global embedding) and the embedding of the last session item (called the local embedding).

\subsubsection{GRU-based graph neural network}
We use a GRU unit instead of LSTM because for session based recommendation models GRUs have been found to be more effective than LSTM units\cite{hidasi2015session}. Let  $G= (V,E)$ represent the batched item sequence graph, where $V$ represents a set of nodes, and $E$ represents a set of edges. Each node $ v \in V$ has a feature vector $h_{v}^{l}$ ($l$ is the layer) associated with it, which is passed as item embeddings into the GNN layer. The GNN layer employs a GRU unit, which allows for sequential processing of node features while maintaining the order of message passing in the graph. We can divide the GRU unit's  task into 3 parts showing how we update embeddings of $v$:\\ 
\textbf{1. Message Passing:} Each node $v$ updates its feature vector $h_v$ from its neighbor $u$ through layer $l$ using a binary message passing function ($f_{msg}^{l}$).
\begin{equation}
f_{msg}^{l} = m_{u^{l} \rightarrow v^{l}}
\label{eq:1}
\end{equation}
\\
\textbf{2. Message Aggregation:} For each node $v$, the received messages from neighbors are aggregated using a GRU unit. $\mathcal{N}_{(v)}$ denotes the set of neighbors of node $v$, and the aggregate message $h_{\mathcal{N}_{(v)}}$ can be represented as:
\begin{equation}
\label{eq:2}
h_{\mathcal{N}_{(v^l)}} = GRU(f_{msg}^{l}, \forall u \in \mathcal{N}_{(v)})
\end{equation}
\\
\textbf{3. Node Update:} Here we combine the node’s own features with the aggregated features from its neighbors and update embeddings for $v$. $\mathbf{W}_{v}$ and $\mathbf{W}_{neigh}$ are weight matrices for transforming the node's own features and the aggregated neighbor features, respectively.


\begin{equation}
\label{eq:3}
h_{v}^{l+1} = \mathbf{W}_{v^l}h_{v^l} + \mathbf{W}_{neigh}h_{\mathcal{N}_{(v)}}
\end{equation}

\subsubsection{Attention mechanism on graphs}
We also leverage an attention mechanism to dynamically weigh node features.  Given an input feature matrix \( X \in \mathbb{R}^{V \times F} \) for a graph with \( V \) nodes and \( F \) features per node, we first apply batch normalization and dropout to improve our model generalization. The implementation then applies linear layers for query, key and value transformations to project the input features to a higher-dimensional space.


\begin{equation}
\label{eq:4}
Q = X \mathbf{W}_Q
\end{equation}


\begin{equation}
\label{eq:5}
K = X \mathbf{W}_K
\end{equation}

\begin{equation}
\label{eq:6}
V = X \mathbf{W}_V
\end{equation}

Here, \( \mathbf{W}_Q \in \mathbb{R}^{F \times H} \), \( \mathbf{W}_K \in \mathbb{R}^{F \times H} \), \( \mathbf{W}_V \in \mathbb{R}^{F \times O} \) are the learnable weight matrices. \( Q \), \( K \), and \( V \) represent the queries, keys, and values, respectively.

Then, edge features \( e \) are computed as the element-wise sum of queries and keys. These features are then passed through a sigmoid function and transformed to derive the attention coefficients,

\begin{equation}
\label{eq:7}
e_{ij} = \sigma(Q_i + K_j) \mathbf{W}_e,
\end{equation}

where \( e_{ij} \) is the attention score between node \( i \) and node \( j \), \( \mathbf{W}_e \in \mathbb{R}^{H} \) is the weight vector for edge features, and \( \sigma \) is the sigmoid activation function.

We then obtain the attention weights by applying a softmax function to the edge attention scores,

\begin{equation}
\label{eq:8}
\alpha_{ij}^{l} = \frac{\exp(e_{ij})}{\sum_{k \in \mathcal{N}(i)} \exp(e_{ik})},
\end{equation}

where \( \mathcal{N}(i) \) denotes the set of neighbors of node \( i \) in the session graph.

Finally, the output representation for each node $i$ is computed as a weighted sum of the value vectors of its neighbors.

\begin{equation}
\label{eq:9}
\mathbf{h}_{i}^{l+1} = \sum_{j \in \mathcal{N}(i)} \alpha_{ij}^{l} \mathbf{V}_j
\end{equation}

\subsubsection{Readout layer}
The Readout module aggregates node features in a sequence using the attention mechanism. This module is particularly suited for scenarios where the importance of nodes varies and needs to be dynamically weighted, making it especially effective for tasks that require feature aggregation.
In our case after item embeddings are generated, the readout technique is used to generate a session embedding. The session is represented as an embedding vector \( s \in \mathbb{R}^{d} \). We define local session embedding (\( s_{local} \)) as the item embedding of the most recent item in the session; for session \( [v_1, v_2, v_3, ..., v_n] \), \( s_{local} = v_n\). Then, we consider the notion of a global session embedding (\( s_{global} \)) which is calculated using the attention mechanism. For the above session example


\begin{equation}
\label{eq:8}
s_{local} = v_n
\end{equation}


\begin{equation}
\label{eq:9}
\alpha_i = q^T\sigma(\mathbf{W}_{1}x_i^{(L)} + \mathbf{W}_{2}x_{n}^{(L)} + r)
\end{equation}


\begin{equation}
\label{eq:10}
s_{global} = \sum_{i=1}^{n} \alpha_i x_{n}
\end{equation}

where \( q \in \mathbb{R}^{d} \) and \( \mathbf{W}_1, \mathbf{W}_2 \in \mathbb{R}^{dxd} \) are parameters that are learned. $r$ is the bias term. $\alpha_i$ denotes the attention score for every item $i$ in the session containing $n$ items. Finally, session embeddings are generated by applying a linear transformation after concatenating the local and global session embedding,

\begin{equation}
\label{eq:11}
s = \mathbf{W}_3 [s_{local} || s_{global}],
\end{equation}

where  \( \mathbf{W}_3 \in \mathbb{R}^{dx2d} \) is a parameter to be learned.

We obtain a score for the $n^{th}$ item using its item embeddings $h_n$ as dependent on the session embeddings using the below equation.

\begin{equation}
\label{eq:12}
h_n = s^T V_n
\end{equation}

As we have set up our model as a multi-item classification problem, we use softmax function that generates a probability distribution of the next item prediction.

\begin{equation}
\label{eq:13}
\hat{y} = softmax(h_n)
\end{equation}

where \( \hat{y} \in \mathbb{R}^{m} \) where \( m \) is number of items. We then use cross-entropy as the loss function.

\begin{equation}
\label{eq:14}
\mathcal{L}(y, \hat{y}) = -  y^T \log(\hat{y})
\end{equation}

All parameters, including the item embeddings, are randomly initialized and jointly learned through an end-to-end backpropagation training process.

The primary distinction between our attention module and the Graph Attention Network (GAT) model\cite{velivckovic2017graph} lies in how connections between items are represented. Our attention module models sequences as a multigraph, meaning an edge exists from \textit{item1} to \textit{item2} only if \textit{item2} appears after \textit{item1} anywhere within a sequence. In contrast, GAT uses a standard adjacency matrix (or edge list), where an edge from \textit{item1} to \textit{item2} exists only if \textit{item2} is the immediate successor of \textit{item1} in the sequence.

\subsection{Real time inferencing}
\textbf{Motivation:} Our original latency test was run on a microservice fetching real-time recommendations in the dev environment. This test used a model trained only on guest actions on frequent items (separate from the one we used for offline evaluation in this paper). Inference latency was found to be over 400 ms, which was unacceptable. This served as the motivation for implementing the nearest neighbor matrix solution for real-time inference.

Inference in our context means that as the guest session changes/evolves dynamically, the model calculates a real-time session embedding and uses it to generate a handful of relevant recommendations from a catalogue of millions of items. In order to meet the latency threshold, we used a heuristic approach of generating a nearest neighbor matrix as part of model training, which has a size of $no \textunderscore items$ x $no \textunderscore neighbors$. We generate this matrix during model training by finding the top-k nearest item embeddings to every item. Suppose we choose $no \textunderscore neighbors = 100$, it means that while inferring recommendations for the session $d \rightarrow e \rightarrow f$, our lookup space for recommendations would not be the entire item catalog but $ \sim $ 300 items (i.e combining 100 pre-calculated neighbors for each of the items $d, e, f$). We use this heuristic approach to address the latency and accuracy trade-off. Figure \ref{fig:inf_architecture} shows how only the the nearest neighbors of session items are considered to generate the final recommendations list.

Approximate Nearest Neighbor (ANN) is a popular method for nearest neighbor generation; however, we opted for a Nearest Neighbor Matrix because ANN-based recommendations across the entire catalog often lacked relevance. This approach allows us to filter the item search space for each item more effectively. Additionally, ANN indexing introduces extra latency, which hindered our microservice from meeting SLA requirements during model inference.

\begin{figure}[H]
\includegraphics[width=\linewidth]{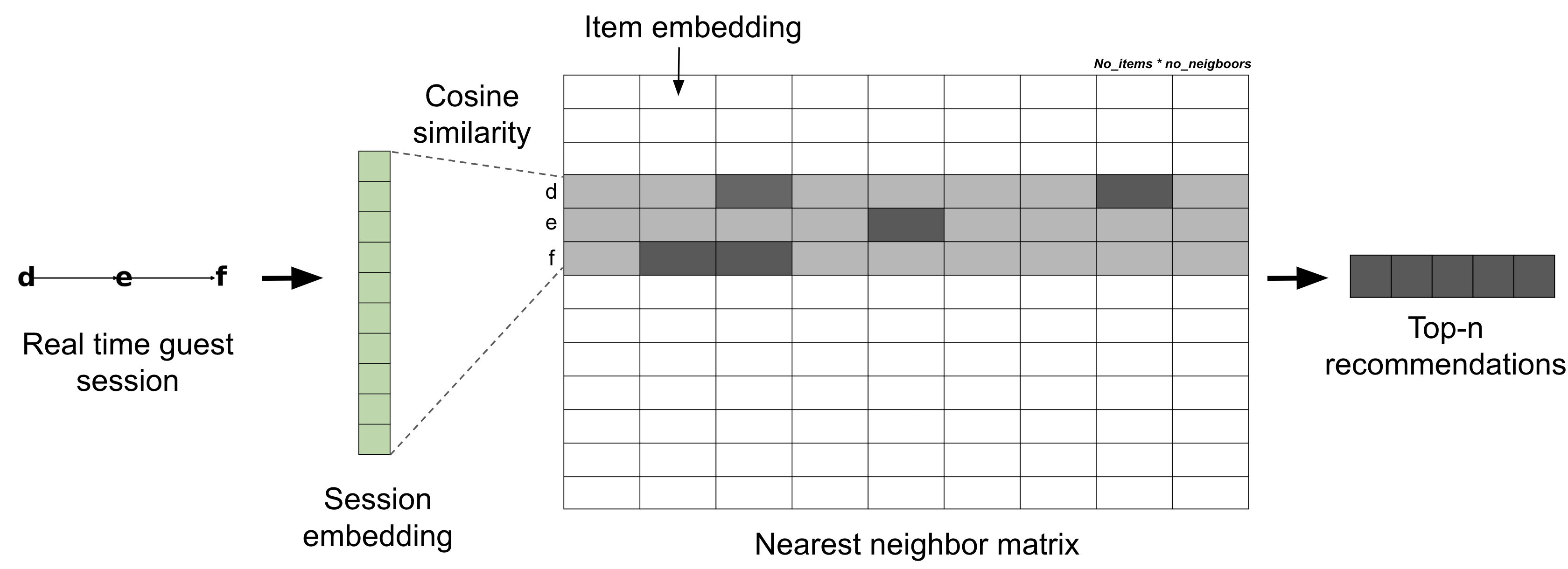}
\caption{Architecture of model inferencing}
\label{fig:inf_architecture}
\end{figure}


\section{Comparison of GRAINRec and LESSR}


The proposed GRAINRec architecture is an enhanced version of the LESSR model \cite{chen2020handling}, incorporating several key modifications to improve performance (Table \ref{tab:model_evaluation}). Unlike the original LESSR model, which included self-loops allowing consecutive item repetitions, GRAINRec was trained without self-loops, which we found improved performance. Cross-category items were excluded from sequences before training and during inference to ensure more coherent session-based recommendations. We also conducted extensive hyperparameter tuning to optimize GRAINRec's performance, an effort not explored in the LESSR study. Finally, to facilitate deployment in Target's production environment, we implemented real-time inference using a nearest neighbor matrix.

\section{Experiments}

\subsection{Evaluation against baseline models}
We compared our proposed model against the following baseline models:\\
1. Item-KNN\cite{davidson2010youtube}: This model recommends similar items to the most recently added item in the session by calculating their cosine similarity scores. Embeddings for items are generated using word2vec.\\
2. FPMC\cite{rendle2010factorizing}: This is a Markov chain-based model for next basket recommendation. For our case we customize it to predict the next item.\\
3. GRU4Rec\cite{hidasi2015session}: This is a recurrent neural network model for session-based next item recommendations.\\
4. SR-GNN\cite{wu2019session}: SR-GNN is a session-based recommendation model using graph neural networks to model session sequences as directed graph-structured data.\\
5. SASRec\cite{kang2018self}: SASRec is a self-attention based sequential model for item recommendations.\\
6. LESSR\cite{chen2020handling}: LESSR model combines graph neural networks with attention.\\
7. GRAINRec: A graph- and attention-based model described above.

\subsection{Dataset statistics and parameter setup}


\begin{table}[H]
    \centering
    \caption{Dataset statistics}
    \label{tab:cat_stats}
    \renewcommand{\arraystretch}{1.2} 
    \setlength{\tabcolsep}{12pt} 
    \begin{tabular}{|c|c|c|}
        \hline
        \textbf{} & \textbf{No. of items} & \textbf{No. of sequences} \\
        \hline
        Frequency items & 36,669 & 68,923,237\\
        \hline
        Discretionary items & 257,397 & 97,140,831 \\
        \hline
    \end{tabular}
\end{table}

All models were trained using item sequences collected by Target, a large retail company. To mitigate data sparsity, we included only items that appeared in the sequence dataset at least 10 times during this period. We capped each sequence length at 20 items and filtered out consecutive duplicate items. The entire catalog was divided into two types: frequency (items usually purchased multiple times, such as groceries, cleaning products, etc) and discretionary (items typically purchased once, such as electronics and furniture). Table \ref{tab:cat_stats} summarizes item and sequence count statistics for both item types used for offline evaluation. While frequency items make up a little over 10\% of the catalog, they account for just over 40\% of the guest action sequences. Also for every sequence of length greater than 2 we also include its prefixes\footnote{prefixes are contiguous subsequences of a given sequence that start at the first element and extend to every other element in the sequence. Ex: For sequence $a \rightarrow b \rightarrow c \rightarrow d$, prefixes are $a  \rightarrow b$, $a \rightarrow b \rightarrow c $ \& $a \rightarrow b \rightarrow c \rightarrow d$.}  for model training.\\
\textbf{Training parameters:} Hyperparameter tuning was performed using the scikit-optimize python library. The  model parameters were set as follows: an embedding dimension of 256, two layers (\textit{GNN-Attention}), dropout rate of 0.146, learning rate of 0.00045, batch size of 1024, and decay rate of 0.0001. Weight matrix dimensions for outputs are as follows: Model Input : $no\_items * emb\_dim$, GNN layer output: $emb\_dim * emb\_dim$, Attention layer output: $1 * emb\_dim$, ReadOut layer: $emb\_dim * (3 x emb\_dim)$, Fully connected layer: $emb\_dim * batch\_size$. We selected a 2-layer model to leverage the strengths of both the GNN and attention mechanisms while maintaining a balance between model complexity and efficiency. Adding more layers would increase the model's complexity, potentially extending training and inference times, which could hinder the model's ability to meet the required service-level agreements (SLAs) for real-time inference. By using a 2-layer architecture, we ensure that the workflow runs more frequently. The model was trained using NVIDIA A100 3G 40GB (3 instances, 40 GB of GPU memory) multi-instance GPUs. \\
\textbf{Inferencing parameters:} For model inference we set an upper limit on the session length to 3. 

For both training and inference, a filter was set on the session to ensure all items in the session belong to the same category as the most recently added item in the session. For example, if the session contains the items: $sugar \rightarrow  iphone \rightarrow  carrots \rightarrow  milk$, the filtered session will be $sugar \rightarrow  carrots \rightarrow  milk$.  $iphone$ belongs to a different category than milk and is excluded from the session, while $sugar$ and $carrot$ belongs to same category as $milk$ and are retained.

\subsection{Inferencing setup}
GRAINRec, like other deployed recommendation models in Target, is a guest-facing application built on an application layer called Kubernetes (k8s). When setting up the inference infrastructure, we faced a critical decision: whether to run inference on CPU or GPU. For our specific use case, CPU was chosen due to its simplicity and favorable cost economics. While GPUs are generally faster for model training, the simplicity of using CPUs for real-time request-response operations without the overhead of batching made it the preferred choice. Additionally, after optimizing our systems, we found that the cost-per-inference on CPU was significantly more economical.

It took less than 50ms (95th percentile latency) to generate GRAINRec recommendations for guests. We were able to achieve this low latency by implementing custom load balancing using gRPC as our RPC framework. We further reduced latency by building an in-memory customized local cache. 
Each instance of the GRAINRec PyTorch model is hosted within a Kubernetes microservice allocated with 2 CPUs and 10 GB of memory. This microservice has direct access to the real-time feature store, which contains comprehensive data on the purchase/browse/add-to-cart history of each guest. The throughput of each k8s pod is around 60 requests per second.


In our k8s environment, we faced challenges scaling GRAINRec PyTorch model inference on CPUs due to resource contention. PyTorch's default behavior of using multiple CPU cores per inference caused performance issues when multiple pods competed for the same cores. To resolve this, we limited each pod to a single thread for PyTorch inference by setting $torch.set\_num\_threads(1)$, ensuring efficient concurrent inference across multiple pods.

\subsection{A/B test results (Online evaluation)}

\begin{table}[H]
    \centering
    \caption{Effect of hyperparameter tuning on online evaluation (CTR- Click Through Rate, AD- Attributable Demand)}
    \label{tab:hyp_param}
    \renewcommand{\arraystretch}{1.5} 
    \setlength{\tabcolsep}{12pt} 
    \begin{tabular}{|c|c|c|}
        \hline
        \textbf{} & \textbf{ CTR} & \textbf{ AD} \\
        \hline
        Before hyperparameter tuning & 6.4 \% $\uparrow$ & 5.8 \% $\uparrow$\\
        \hline
        After hyperparameter tuning & 10.1 \% $\uparrow$ & 9.2 \% $\uparrow$ \\
        \hline
    \end{tabular}
\end{table}

The proposed model was A/B tested against a link prediction model already running in production. Increase in metrics is compared against the link prediction model. The evaluation was done against two metrics:\\
\textbf{1. Click Through Rate (CTR)}: This metric quantifies how often guests click on recommended items compared to how often those items are displayed to guests.\\
\textbf{2. Attributable Demand (AD)}: This metric quantifies the \$ sales generated directly from the guests purchasing the recommended items.

Table \ref{tab:hyp_param} presents the impact of hyperparameter tuning on key performance metrics for an item recommendation system. Before hyperparameter tuning, the CTR increased by 6.4\% and AD increased by 5.8\%. When A/B testing was done after  tuning, CTR rose to 10\%, and AD increased to 9\%. These results demonstrate that hyperparameter tuning significantly enhances the effectiveness of the recommendation system, leading to greater user engagement.

\subsection{Evaluation on dataset (Offline evaluation)}

\begin{table*}
    \centering
    \caption{Evaluation of proposed approach (GRAINRec) against baseline models}
    \label{tab:model_evaluation}
    \renewcommand{\arraystretch}{1.2} 
    \setlength{\tabcolsep}{4pt} 
    \begin{tabular}{|p{0.12\textwidth}|p{0.08\textwidth}|p{0.08\textwidth}|p{0.08\textwidth}|p{0.08\textwidth}|p{0.08\textwidth}|p{0.08\textwidth}|p{0.08\textwidth}|p{0.08\textwidth}|p{0.08\textwidth}|}
        \hline
        \multirow{2}{*}{\textbf{Models}} & \multicolumn{3}{c|}{\textbf{Frequency items }} & \multicolumn{3}{c|}{\textbf{Discretionary items}} & \multicolumn{3}{c|}{\textbf{Entire catalog}} \\
        \cline{2-10}
        & \textbf{hit@10} & \textbf{mrr@10} & \textbf{ndcg@10} & \textbf{hit@10} & \textbf{mrr@10} & \textbf{ndcg@10} & \textbf{hit@10} & \textbf{mrr@10} & \textbf{ndcg@10} \\
        \hline
        Item-KNN & 0.134 & 0.127 & 0.131 & 0.111 & 0.105 & 0.108 & 0.122 & 0.118 & 0.12 \\
        \hline
        FPMC & 0.151 & 0.143 & 0.149 & 0.123 & 0.116 & 0.119 & 0.145 & 0.133 & 0.139 \\
        \hline
        GRU4Rec & 0.219 & 0.103 & 0.121 & 0.207 & 0.101 & 0.123 & 0.221 & 0.123 & 0.129 \\
        \hline
        SR-GNN & 0.222 & 0.101 & 0.123 & 0.209 & 0.105 & 0.122 & 0.231 & 0.109 & 0.133 \\
        \hline
        SASRec & 0.248 & 0.119 & 0.148 & 0.228 & 0.109 & 0.139 & 0.235 & 0.134 & 0.144 \\
        \hline
        \hline
        LESSR & 0.255 & 0.122 & 0.151 & 0.234 & 0.112 & 0.141 & 0.239 & 0.136 & 0.147 \\
        \hline
        GRAINRec & \textbf{0.258} & \textbf{0.124} & \textbf{0.153} & \textbf{0.238} & \textbf{0.114} & \textbf{0.144} & \textbf{0.245} & \textbf{0.139} & \textbf{0.149} \\
        \hline
        \hline
        Improvement & 1.18 \% & 1.64\% & 1.32\% & 1.71\% & 1.79\% & 1.98\% & 2.51\% & 2.21\% & 1.36\% \\
        \hline
    \end{tabular}
\end{table*}

We evaluate our model against baselines listed in Section VI. A. using 3 popular metrics:\\
\textbf{1. Hit rate at 10 (hit@10)}: It indicates the proportion of times a relevant item is found within the top 10 recommendations provided by the system.\\
\textbf{2. Mean Reciprocal Rank at 10 (mrr@10)}: It measures the average reciprocal rank (average of inverse of the ranks) of the first relevant item across multiple test cases.\\
\textbf{3. Normalized Discounted Cumulative Gain at 10 (ndcg@10)}: It combines the concepts of cumulative gain, discounting the gain of relevant items based on their position in the recommendation list, and normalizing the result to ensure comparability across different recommendation lists.

The performance analysis of various recommendation models across different item categories is encapsulated in Table \ref{tab:model_evaluation}.

For frequency item categories, Item-KNN is the least performing model with scores of 0.134 (hit@10), 0.127 (mrr@10), and 0.131 (ndcg@10). As a word2vec based approach, its results are low due to its inability to capture any patterns in the item sequences. FPMC showed slight improvement over Item-KNN with scores of 0.151 (hit@10), 0.143 (mrr@10), and 0.149 (ndcg@10) as it captures position information by modelling sequential data through Markov chains. GRU4Rec and SR-GNN significantly outperformed the traditional models as they employ deep learning techniques. GRU4Rec, with 0.219 (hit@10), 0.103 (mrr@10), and 0.121 (ndcg@10) leverages gated recurrent units to capture sequential dependencies. Its improved performance can be attributed to its gated neural network model that stores information from previous item. A small drawback of the model is that information passing happens only in one dimension, which is overcome by graphical representation of sequences. SR-GNN with 0.222 (hit@10), 0.101 (mrr@10), and 0.123 (ndcg@10) further enhances this by incorporating graph neural networks to model complex item relationships better. SASRec exhibited stronger performance with 0.248 (hit@10), 0.119 (mrr@10), and 0.148 (ndcg@10). Its self-attentive mechanism effectively models sequence patterns that captures long range dependencies, providing a significant boost in recommendation accuracy. LESSR model shows performance of 0.255 (hit@10), 0.124 (mrr@10) and 0.151 (ndcg@10) which is better than the previous baselines as it incorporates both GNN and attention. Our proposed GRAINRec emerged as the best performer in this category with 0.258 (hit@10), 0.124 (mrr@10), and 0.153 (ndcg@10). In addition to the multi-layers attention and GNN based architecture, what contributes to the performance improvement is removal of self-loops, cross category dependency and nearest neighbor matrix.

In discretionary item categories, the trends observed is similar as in frequency item categories. Item-KNN and FPMC maintain their lower performance, with FPMC slightly better at 0.123 (hit@10), 0.116 (mrr@10), and 0.119 (ndcg@10) compared to Item-KNN's 0.111 (hit@10), 0.105 (mrr@10), and 0.108 (ndcg@10). GRU4Rec (0.207 hit@10, 0.101 mrr@10, 0.123 ndcg@10) and SR-GNN (0.209 hit@10, 0.105 mrr@10, 0.122 ndcg@10) continue to show their strength.  SASRec again show superior performance, achieving 0.228 (hit@10), 0.109 (mrr@10), and 0.139 (ndcg@10). LESSR shows even better performance with 0.234 (hit@10), 0.112 (mrr@10), and 0.141 (ndcg@10). GRAINRec leads with 0.238 (hit@10), 0.114 (mrr@10), and 0.144 (ndcg@10), indicating its robustness.

When considering the entire catalog, Item-KNN (0.122 hit@10, 0.118 mrr@10, 0.12 ndcg@10) and FPMC (0.145 hit@10, 0.133 mrr@10, 0.139 ndcg@10) again exhibit lower scores compared to more sophisticated models. GRU4Rec and SR-GNN continue their competitive performance, with GRU4Rec at 0.221 (hit@10), 0.123 (mrr@10), and 0.129 (ndcg@10) showing strong consistency across metrics. SR-GNN, however, shows a marked drop in mrr@10 (0.109), possibly indicating some limitations in its architecture for broader item catalogs despite having good hit@10 (0.231) and ndcg@10 (0.133) scores. SASRec remains effective with scores of 0.235 (hit@10), 0.134 (mrr@10), and 0.144 (ndcg@10). LESSR outperforms other baselines with 0.239 (hit@10), 0.136 (mrr@10), and 0.147 (ndcg@10), but GRAINRec maintains its leading position, achieving 0.245 (hit@10), 0.139 (mrr@10), and 0.149 (ndcg@10).

Our proposed model shows an average improvement of over 1.5\% over LESSR for all metrics. Comparing the performances across frequency, discretionary, and entire catalog categories we observe that the model trained on frequency categories performs the best. This can be attributed to the lower data sparsity in the frequency dataset, which allows the model to learn item relationships more effectively. On the other hand, items in the discretionary category are more numerous but occur less frequently, leading to poorer learning of their representations.

\subsection{Multi-class classification}
\begin{table}[H]
    \centering
    \caption{Comparison of single- and multi-class classification for GRAINRec}
    \label{tab:classification_result}
    \renewcommand{\arraystretch}{1.3} 
    \setlength{\tabcolsep}{6pt} 
    \begin{tabular}{|p{4cm}|c|c|c|}
        \hline
        \textbf{} & \textbf{hit@10} & \textbf{mrr@10} & \textbf{ndcg@10} \\
        \hline
        Single-class classification & 0.245 & 0.139 & 0.149\\
        \hline
        Multi-class classification & 0.156 & 0.091 & 0.101\\
        \hline
    \end{tabular}
\end{table}

Table \ref{tab:classification_result} compares the performance of GRAINRec under the single-class and multi-class classification settings. Single-class classification significantly outperforms multi-class classification across all evaluation metrics. Specifically, hit@10, mrr@10, and ndcg@10 values for multi-class classification show a drop of 36.33\%, 34.53\% and 32.21\% respectively, compared to single-class classification. This could be attributed to the fact that as the number of classes grows (the catalog has roughly 600k items), the number of decision boundaries that a learning algorithm must address also increases. Experimental evidence\cite{del2022multiclass} suggests that with more decision boundaries, the complexity of the problem rises which significantly reduces model performance.

\section{Ablation Studies}

In this section, we conduct a series of ablation studies to systematically investigate the contributions of various components of our proposed model to its performance of ndcg@10 metric. By selectively removing/modifying specific parameters of the architecture, we aim to understand the impact of each component on the model's performance, providing deeper insights into the strengths and potential areas for improvement of our approach.
\subsection{Ordering of layers}

\begin{figure}
\includegraphics[width=\linewidth]{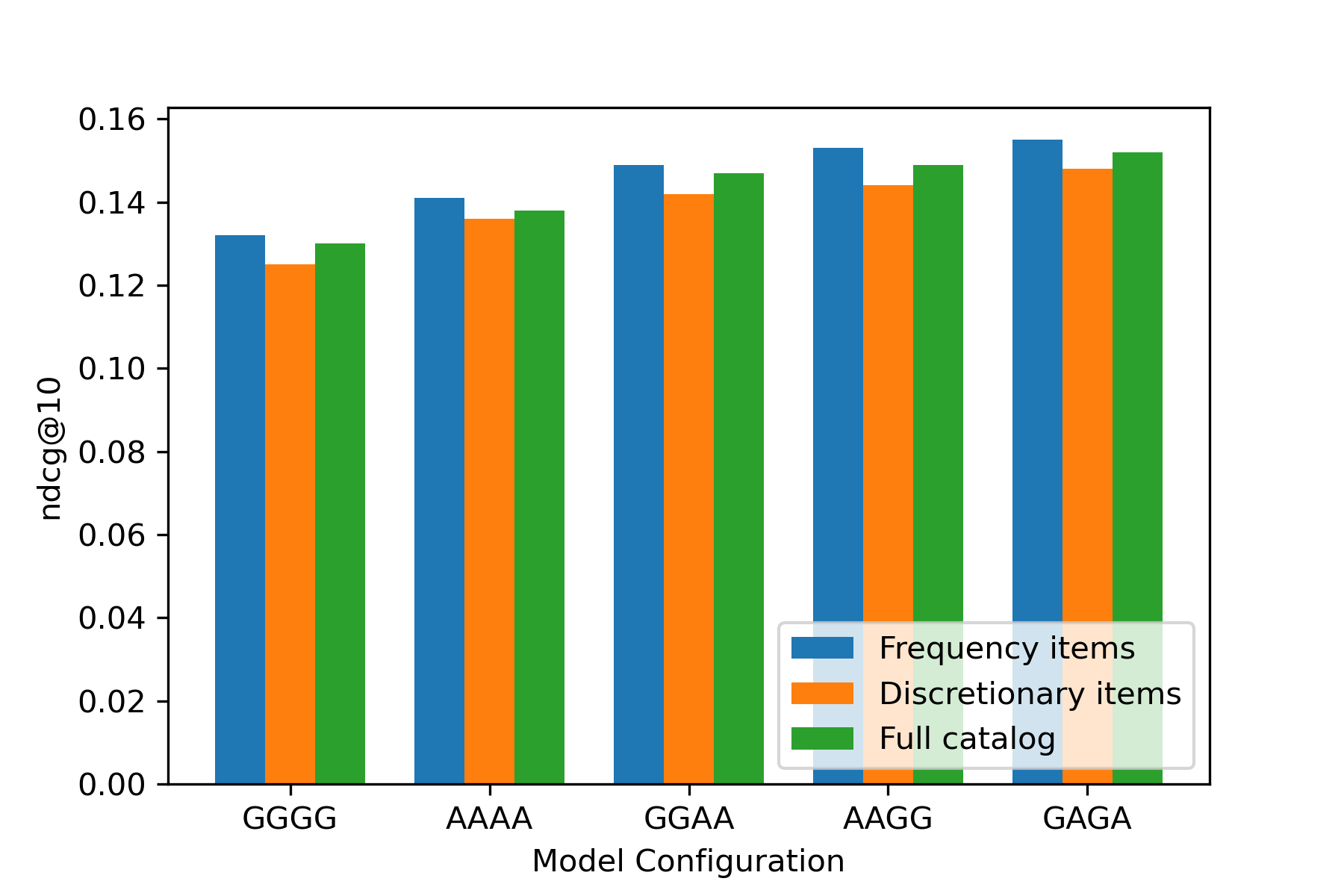}
\caption{ndcg@10 for different 4-layer configurations (G- Graph neural network, A- Attention mechanism)}
\label{fig:ndcg_layers}
\end{figure}

Figure \ref{fig:ndcg_layers} summarizes the performance of the proposed model across different item categories using various configurations of graph neural network (G) and attention (A) layers. The configurations tested include a 4-layer GNN model (GGGG), a 4-layer attention model (AAAA), a mixed 2-layer GNN followed by 2-layer attention model (GGAA), the reverse order (AAGG), and an alternating layer model (GAGA).

The results show that GAGA, the proposed alternating layer model, outperforms all other configurations, demonstrating the effectiveness of alternating GNN and attention layers. This approach effectively leverages the strengths of both types of layers, with the GNN layers capturing the context of closer items and the attention layers capturing dependencies of more distant items. This layered alternation helps mitigate the \textit{over-squashing}\cite{alon2020bottleneck} issue seen in continuous GNN layers, where information propagation across long distances in the graph becomes inefficient. The AGAG configuration showed similar performance to GAGA, reinforcing the advantage of alternating the two types of layers for improving recommendation accuracy across all item categories.

\subsection{Size of nearest neighbor matrix}

\begin{figure}
\includegraphics[width=\linewidth]{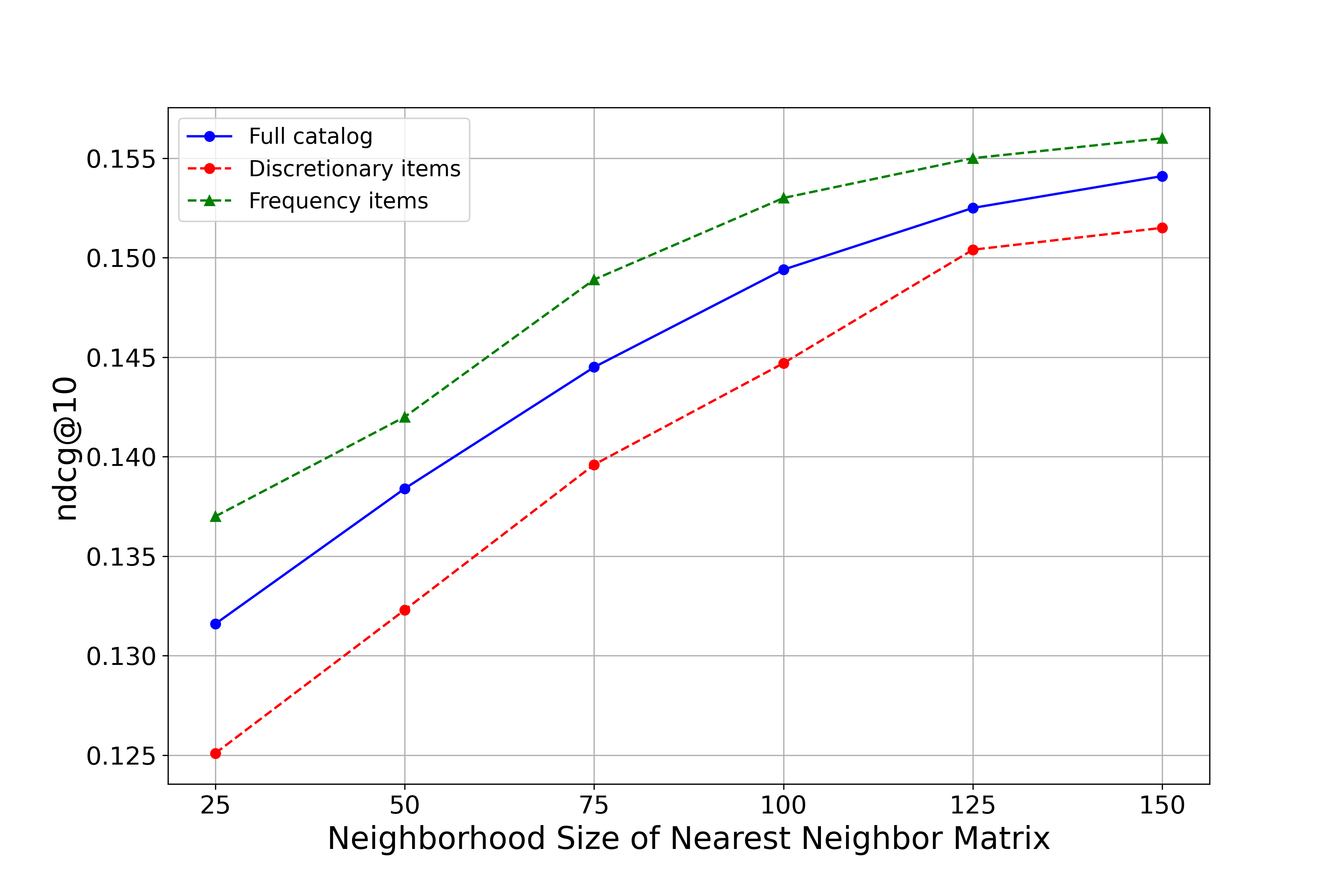}
\caption{ndcg@10 for different neighborhood sizes in the Nearest Neighbor Matrix}
\label{fig:ndcg_neighborhood}
\end{figure}

We investigate the relationship between model performance and the size of the nearest neighbor matrix in Figure \ref{fig:ndcg_neighborhood}, evaluating the model using ndcg@10 across six different neighborhood sizes: 25, 50, 75, 100, 125, and 150. For the full catalog, the ndcg@10 value increases from 0.1316 to 0.1541. Discretionary items show a similar improvement, rising from 0.1251 to 0.1515, while frequency items exhibit the most significant growth, from 0.1370 to 0.1560. These results suggest that larger neighborhood sizes enhance recommendation accuracy across all categories, with frequency items achieving the highest ndcg@10 values. While the general trend shows increasing ndcg@10 values, with larger neighborhood sizes we observe diminishing marginal returns beyond size of 100. Specifically, the rate of increase in ndcg@10 values slows down past this point. Consequently, we select a neighborhood size of 100 for our model, as larger sizes lead to inferencing times that fail to meet platform's SLA.

\subsection{Embedding dimension and number of layers}

\begin{figure*}[h]
\includegraphics[width=\linewidth]{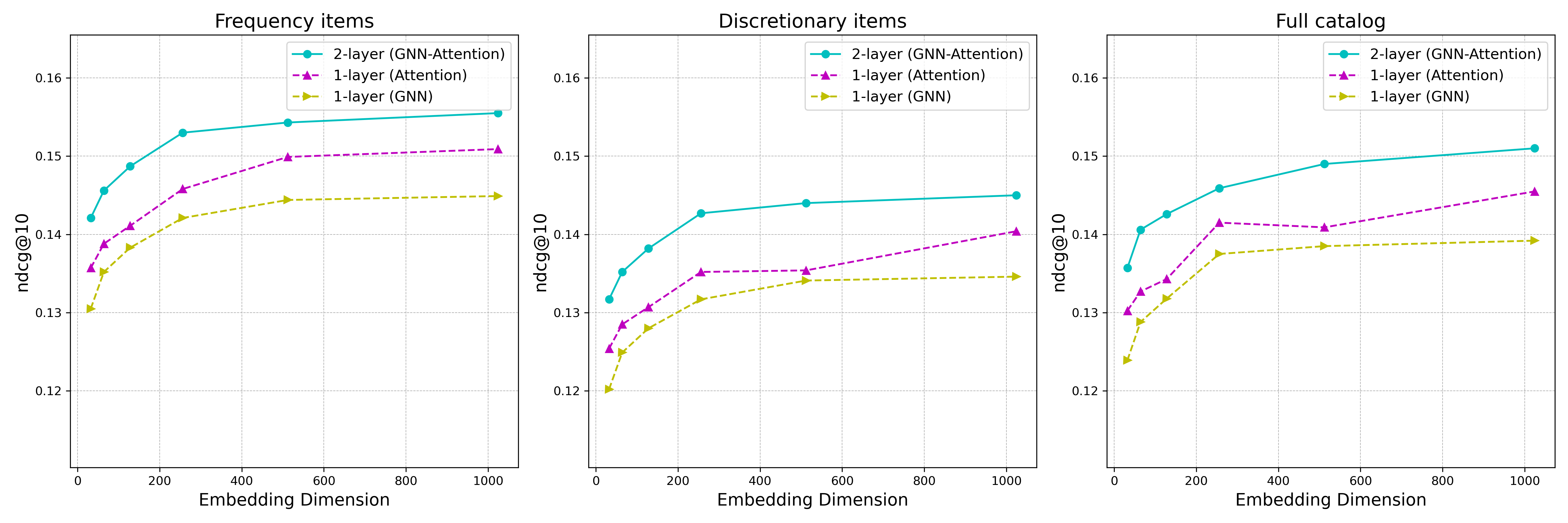}
\caption{ndcg@10 for different neighborhood sizes in the nearest neighbor matrix}
\label{fig:ndcg_emb}
\end{figure*}

The experiment results are presented in Figure \ref{fig:ndcg_emb}, which displays the performance of model across varying embedding dimensions: 32, 64, 128, 256, 512, and 1024. The models compared are a 2-layer Graph Neural Network with Attention (GNN-Attention), a 1-layer Attention model, and a 1-layer Graph Neural Network. Each subplot in the figure corresponds to a different item category: frequency items, discretionary items, and full catalog. All three subplots have a consistent y-axis range, making it easier to compare the results across different item categories.

In the first subplot, representing frequency items, the 2-layer GNN-Attention model consistently outperforms the other models across all embedding dimensions. Specifically, the ndcg@10 values for the GNN-Attention model range from 0.1421 to 0.1555. The 1-layer Attention model shows moderate performance, with ndcg@10 values ranging from 0.1357 to 0.1509. The 1-layer GNN model demonstrates the lowest performance, with ndcg@10 values between 0.1305 and 0.1449. This indicates that incorporating both GNN and Attention mechanisms yields the best results for frequency item recommendations.

The second subplot, which depicts discretionary items, shows a similar trend. The 2-layer GNN-Attention model again leads in performance, with ndcg@10 values between 0.1317 and 0.1450. The 1-layer Attention model follows, with values ranging from 0.1254 to 0.1404. The 1-layer GNN model has the lowest ndcg@10 values, ranging from 0.1202 to 0.1346. 

In the third subplot, which represents the full catalog, the performance of the models is evaluated on all items. The 2-layer GNN-Attention model maintains its superior performance, with ndcg@10 values ranging from 0.1357 to 0.1510. The 1-layer Attention model achieves ndcg@10 values between 0.13024 and 0.14549. The 1-layer GNN model shows the lowest performance with values ranging from 0.1239 to 0.1392. The consistent outperformance of the 2-layer GNN-Attention model across different item categories and embedding dimensions highlights its robustness and suitability for diverse recommendation scenarios.

Overall, the experiment results clearly demonstrate the effectiveness of the proposed 2-layer GNN-Attention model in achieving higher ndcg@10 values compared to the 1-layer Attention and 1-layer GNN models. The enhanced performance is observed across all embedding dimensions and item categories, indicating the model's capability to capture complex interactions and improve recommendation accuracy in real-time session-based systems. While GNNs are adept at capturing short-range dependencies, attention mechanisms additionally captures long-range dependencies, thereby providing a more comprehensive representation of the data. Furthermore, we observed that model performance improves with an increase in embedding dimensions, as larger dimensions enable the model to capture a greater amount of information. However, this improvement follows a trend of diminishing returns, where the incremental gains in performance decrease as the embedding dimension continues to increase.

\section{Conclusion and Future Work}
We propose a session-based recommendation model that integrates graph neural network and attention mechanisms to generate item recommendations. We also integrated a readout layer that uses the attention mechanism to generate session embeddings capturing the context of all items in the session. We proposed a k-nearest neighbor-based heuristic approach that allows recommendations to be generated in real time. We compared our model against SOTA session-based models and saw a improvement across all metrics. The model also showed an improvement in click-through rate and attributable demand during online evaluation.

As part of future work, we will evaluate our model using a contrastive learning setup with InfoNCE loss function. While the proposed model is implemented for the add to cart placement, in future we intend to test it in placements that consider other guest actions such as page views or item purchases. We are currently in the process of setting up and A/B testing said placements. We also intend to work on improving the model's ability to handle cross-category item relationships. Currently we are in the process of integrating item facets (product title, brand, color, size) and language models during model training to generate pertained embeddings that would likely improve the quality of recommendations.\\
\\
\textbf{Acknowledgment:} The authors would like to thank all members of the Item Personalization team at Target for their constructive feedback and support for this work.

\section{COMPANY PORTRAIT}
Target Corporation is a leading American retailer based in Minneapolis, Minnesota, operating around 2,000 stores nationwide with a workforce of about 400,000 employees. Known for its high-quality products and excellent customer service, Target offers a wide range of products, including food, clothing, home goods, electronics, and health and beauty items, with a mission to help families discover the joy of everyday life.


\end{document}